%% file: MICCAI2026-arxiv.tex
\documentclass[runningheads]{llncs}
\usepackage[T1]{fontenc}
\usepackage{amssymb}
\usepackage{amsmath}
\usepackage{pifont}
\newcommand{\cmark}{\ding{51}}%
\newcommand{\xmark}{\ding{55}}%
\usepackage{graphicx,verbatim}
\usepackage{caption}
\usepackage{subcaption}
\usepackage{float}
\usepackage{multirow}
\usepackage{rotating}
\usepackage[export]{adjustbox}
\usepackage{tabularx}
\usepackage{booktabs, makecell}
\usepackage{xcolor}
\usepackage[colorinlistoftodos,textsize=small]{todonotes}
\usepackage{hyperref}

\usepackage{orcidlink}
\usepackage[nolist]{acronym}

\newcommand\blfootnote[1]{%
  \begingroup
  \renewcommand\thefootnote{}\footnote{#1}%
  \addtocounter{footnote}{-1}%
  \endgroup
}

\begin{document}
\input{acronym}

\title{Multi-Camera AR Guidance System for Surgical Instrument Handling and Assembly: Investigating Workload and Efficiency}

\titlerunning{Multi-Camera AR Guidance System for Surgical Instrument Handling}

\author{
Shiyu Li*\inst{1-3}\orcidlink{0000-0002-0888-2496},
Julian Kreimeier*\inst{1-4}\orcidlink{0000-0001-6861-711X},
Hannah Schieber\inst{1-4}\orcidlink{0000-0002-5786-3283}, 
Dirk Müller\inst{2}\orcidlink{0009-0005-7115-9224},
Bernhard Kainz \inst{4,5}\orcidlink{0000-0002-7813-5023}, 
Rüdiger von Eisenhart-Rothe\inst{2,3}\orcidlink{0000-0002-2352-1357}, \and
Daniel Roth\inst{1-3}\orcidlink{0000-0001-5175-1566}}
\authorrunning{S. Li et al.}
%
\institute{Technical University of Munich, \\ Human-Centered-Computing and Extended Reality Lab
\and TUM University Hospital, Orthopedics and Sports Orthopedics
\and Munich Institute of Robotics and Machine Intelligence (MIRMI)
\and FAU Erlangen-Nürnberg, Erlangen, Germany
\and Department of Computing, Imperial College London, London, UK \\
\email{\{shiyu.li,julian.kreimeier,daniel.roth\}@tum.de} \blfootnote{* equal contribution}}
  
\maketitle

\begin{abstract}
The handling and assembly of instruments during surgery imposes high cognitive demands on scrub nurses, particularly when instruments are unfamiliar. We present a supporting guidance system for surgical instrumentation that combines multi-camera 6D pose estimation with augmented reality in-situ visualization on a head-mounted display without the requirement for additional markers. Pose estimation and consecutive camera calibration are achieved through known objects. The 6D pose estimation network is trained purely on synthetic data, aiming for better generalizability and real-world applicability. The AR guidance displays tooltip localization cues and step-wise assembly animations. Via gaze-based selection and a foot pedal, users can switch between assembly steps in intraoperative use.

In a technical evaluation, our approach outperforms state-of-art 6D pose estimation. A user study with 29 scrub nurses was conducted in a surgical simulation of knee arthroplasty, comparing the system against a paper manual. AR guidance significantly reduced the perceived workload compared. Objectively, AR guidance reduced task completion time by 21.3\% (4.76 minutes). Specifically, scrub nurses less experienced with the instrument set benefited when using the system. Error frequencies were comparable between conditions. Qualitative feedback highlighted improved process clarity, reduced information overload, and perceived independence. To summarize, our marker-free multi-camera AR guidance approach for surgical instruments can, subjectively and objectively, improve intraoperative instrumentation performance, particularly for untrained scrub nurses.

\end{abstract}

\section{Introduction and Related Work}
In the \ac{or}, scrub nurses sterilely support surgeons in a sensitive environment under time pressure. Endoprosthetic instruments have become very complex, and manufacturer-dependent. Due to this and a general workforce shortage, unexperienced staff may need to handle unfamiliar instruments \cite{cramer_requirement_2024}. Yet flawless handling remains essential for efficiency and patient safety, as pauses during surgery pose unnecessary risks. 

The WHO's Patient Health and Safety plan \cite{noauthor_global_nodate} requires that operating manuals and safety instructions be available at the point of use, which is not the current gold standard. Currently, when problems occur, PDF or paper manuals are requested, often from outside of the \ac{or}, and reviewed, often by additional personnel. This results in longer interruptions \cite{goras_tasks_2019}. Linear digital manuals or tools for assembly guidance provided by different manufacturers  are not broadly embedded, nor are they context dependent. AR guidance with in-situ instrument detection could serve as a solution to fulfill the WHO criteria. 

\textbf{Augmented Reality Guidance} recently focuses on surgeons' real-time navigation and visualization \cite{acar2025navius}, instrumentation \cite{muller2020augmented}, and registration with deformable tissue \cite{yang2025augmented}. Also, marker-free situs registration is being extensively researched \cite{liebmann2024automatic}. However, prior to the surgeons' activities, handling and assembly of the instruments by the scrub nurse has not been extensively investigated, though this process is often a pain point for inexperienced staff and promising to support by AR guidance \cite{kreimeier_visual_2024}. Such situation-aware \ac{ar} can guide scrub nurses by visualizing necessary information directly onto the surgical trays. Using relevant objects' pose information, the \ac{ar} system can show in-situ tooltips ("pick those") and assembly animations ("assemble them like this") providing step-by-step guidance \cite{kleinbeck_artfm_2022}, reducing cognitive load, errors, and task completion time \cite{blattgerste_-situ_2018}. Yet, empirical insights in an \ac{or} situation are missing.

\textbf{Surgical Instrument Detection and 6D Pose Estimation} is predominantly studied in endoscopic videos \cite{demir2023deep}, while non-endoscopic approaches either rely on fiducial markers to achieve high accuracy \cite{hogenkamp20256d,martin2023sttar} or investigate multi-camera marker-free 6D pose estimation methods with two surgical tools \cite{hein2025next,agethen2025recurrent}. Given this context, a recent approach introduced nine surgical tools being used to perform static and dynamic camera calibration on-the-fly without a calibration board \cite{li2026multicam}. Also, the potential of synthetically-trained deep learning–based approaches is not fully leveraged \cite{schieber_asdf_2024,labbe2020cosypose}. It remains unclear how multi-camera, entirely marker-free pose estimation performs in a full-fidelity surgical simulation to inform an AR guidance system for inexperienced scrub nurses. To address these gaps, we contribute:

\begin{itemize}
    \item An \ac{ar} guidance system for surgical instrument handling assembly connected to deep learning-based 6D pose estimation.
    \item A multi-camera pose estimation method trained on synthetic data, overcoming challenges in data annotation, allowing for generalization to manifold instrument sets.
    \item A technical evaluation of the system's performance.
    \item An expert user study with scrub nurses comparing paper-based manuals and our \ac{ar} guidance system in a simulated \ac{or} scenario.
\end{itemize}

Our work is the first to benchmark multi-camera \ac{ar} assembly guidance in an \ac{or} scenario. Our guided \ac{ar} system improved overall user experience and reduced task completion time by 21.3\% or 4.76 minutes, respectively, in a sample of scrub nurses with different levels of experience.

\begin{figure}[t!]
\centering
  \includegraphics[trim= 20 60 0 10, clip, width=\linewidth]{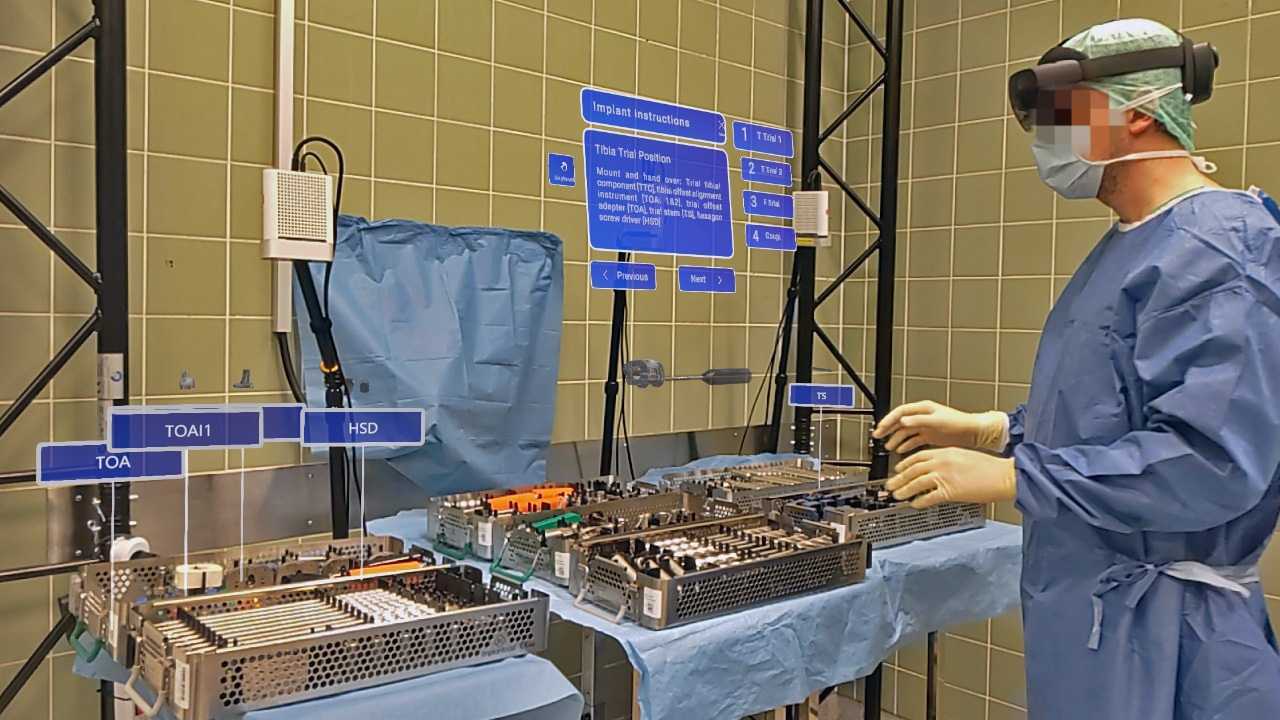}
  \caption{Informed by marker-free multi-camera pose estimation, our AR guidance system visualizes tooltips and an assembly animation for knee arthroplasty.}
  \label{fig:teaser}
\end{figure}
\section{Method}

We propose an end-to-end approach for multi-object, multi-camera 6D pose estimation connected to an in-situ \ac{ar} application building upon user requirements \cite{cramer_requirement_2024}, allowing users to assemble challenging medical instruments.

Our guidance system detects a) surgical trays and the assembly table for relevant parts and b) provides visual guidance by means of \ac{ar} in-situ tool tips and assembly visualizations. The backend runs a hierarchical marker-free multi-camera pose estimation to comply with sterility and space constraints in the \ac{or}, fusing multiple camera streams. Furthermore, we leverage recognized object pose estimation to calibrate the \ac{hmd} with non-overlapping static cameras for maximum coverage. On the \ac{hmd} we run a Unity application developed with MRTK2 highlighting the object locations detected by our pose estimatior, which is trained on synthetic data. 

\subsection{Synthetic Data Generation}

For synthetic data generation, we scan each instrument (DePuy Synthes TFN-ADVANCED™ Proximal Femoral Nailing System and Implantcast GenuX MK implant) and trays with a Shining EinScan structured-light scanner with up to 0.05~mm accuracy. We mitigate artifacts via MeshLab post-processing. The post-processed scans are used in BlenderProc~\cite{denninger_blenderproc_2020} adding domain randomization (varying backgrounds, lighting, camera poses, distractor objects~\cite{schieber_indoor_2024}) to generate training ground truth object poses and images.

\subsection{Hierarchical Multi-Camera Pose Estimation}

\subsubsection{Model Configuration}

Our pose detection model integrates a real-time capable object detector and pose estimator to balance speed and accuracy. We use object and keypoint detection modules within the YOLOX architecture~\cite{yolox2021}. We integrate DDC from RTMO~\cite{lu2023rtmo} to increase keypoint accuracy from regressed \ac{bbox}. Object poses are recovered via RANSAC-based \ac{pnp} from predicted \ac{bbox} and keypoints \cite{schieber_asdf_2024,li_gbot_2024}. 

\begin{figure*}[t]
    \centering
    \includegraphics[width=\textwidth]{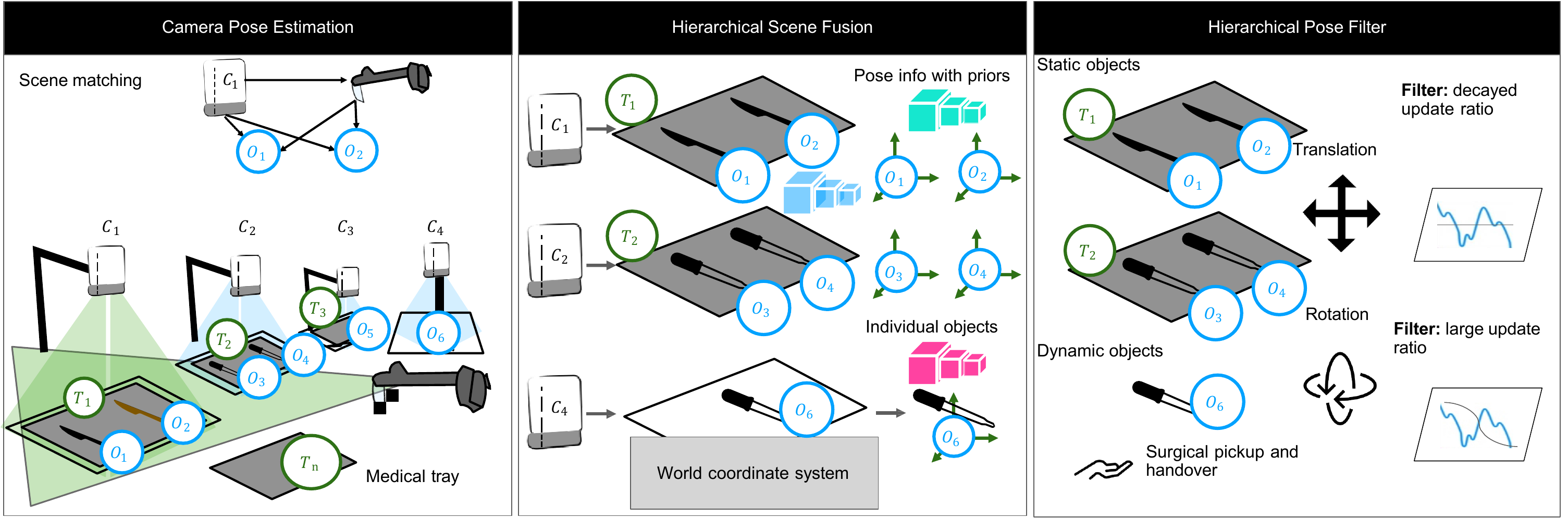}
    \caption{System overview: Four static top-down cameras observe the surgical trays and deposit table (left), with RGB data fused and processed on a server for marker-free pose estimation and camera calibration (center, right).}        
    \label{fig:hardwareSetup}
\end{figure*}

\subsubsection{Training}

For training, we exclusively utilize synthetic images via our data generation. To optimize the \ac{cnn}, we use the 2D MLE, 3D keypoint \cite{hu_wide-depth-range_2021} and proxy loss following MultiCam \cite{li2026multicam}. Hyperparameters selection follows RTMO \cite{lu2023rtmo}.

\subsubsection{Inference}

The model runs the same configuration as during training, but is extended with filtering to optimize the robustness under real-world \ac{or} conditions and reduce pose misalignment \cite{khuong_effectiveness_2014}. We combine the pose estimation of individual instruments and surgical instrument trays using the \ac{hmd} camera and static cameras \cite{li2026multicam}. Instrument poses are first estimated on the surgical instrument table; if unsuccessful, the instrument is assumed to remain in its tray. Since relative poses within trays are predefined, the world pose of an instrument can be computed inside them.

\paragraph{Hierarchical Scene Fusion}

We calculate each camera pose using spatiotemporal overlapping object candidates following \cite{labbe2020cosypose,li2026multicam}. Then, we define our hierarchical scene structure with camera and objects nodes. Given each camera pose, we fuse the object poses from trays and individual objects in each camera view.

\paragraph{Hierarchical Pose Filter}

All pose estimates are processed by a two-stage filter: First, a plausibility check rejects physically implausible poses (e.g., excessive distance, camera proximity, or inverted trays). Second, valid poses are smoothed using an exponential moving average, with linear filtering for translations and spherical linear interpolation for rotations as follows: 

\begin{equation}
\mathbf{t}_{\text{smooth}} = \alpha \mathbf{t}_{\text{curr}} + (1 - \alpha) \mathbf{t}_{\text{prev}}, \text{~~~}
\mathbf{q}_{\text{smooth}} = \text{slerp}(\mathbf{q}_{\text{prev}}, \mathbf{q}_{\text{curr}}, \alpha)
\end{equation}

where $\mathbf{t}_{\text{curr}} \in \mathbf{R}^3$ and $\mathbf{q}_{\text{curr}}$ are the current translation and rotation, $\mathbf{t}_{\text{prev}}$ and $\mathbf{q}_{\text{prev}}$ are the previous ones, and $\alpha\in[0,1]$ is the filter coefficient.

Considering static and dynamic objects in the scene, we apply a hierarchical pose filter using different filter coefficients, i.e. , a fixed coefficient $\alpha=0.2$ for individual dynamic instruments and exponentially decayed coefficient $\alpha = e^{\frac{-i}{300}}$ for static trays to converge to a stable pose, where $i$ is the number of frames. We assume static cameras and trays for the technical evaluation and the user study. To enable using moving trays and cameras, static constraints can be disabled.
 
\subsection{System Architecture}

We use RGB feeds from four Azure Kinect cameras and a Microsoft HoloLens 2 \ac{hmd}, see Fig.~\ref{fig:hardwareSetup}. Our C++ implementation builds upon OpenMMLab \cite{lu2023rtmo,mmpose2020}, achieving an average runtime around 200~ms per frame on a workstation with a RTX 4080 GPU (16 GB VRAM) and an AMD Ryzen 9 9900X CPU with 64~GB RAM. The \ac{hmd} is connected via TCP client-server \cite{dibene2022hololens} using WiFi to the workstation which executes the multi-camera, multi-object 6D pose estimation. The measured end-to-end latency for data streaming from the workstation to the \ac{hmd} is approx. 60~ms. The server analyzes the video streams from four static and one dynamic camera, each with 15 frames per second (tradeoff between computation speed and precision). Our \ac{hmd} object pose data refreshment rate is three seconds.

\section{Technical Evaluation}
\begin{figure*}[t!]
\setlength\tabcolsep{1pt}
\centering
\begin{tabularx}{0.85\textwidth}{l XXXXX }
& \centering HMD view & \centering Static camera 1 & \centering Static camera 4 \cr \\
\rothead{GT}        &   \includegraphics[width=\hsize,valign=m]{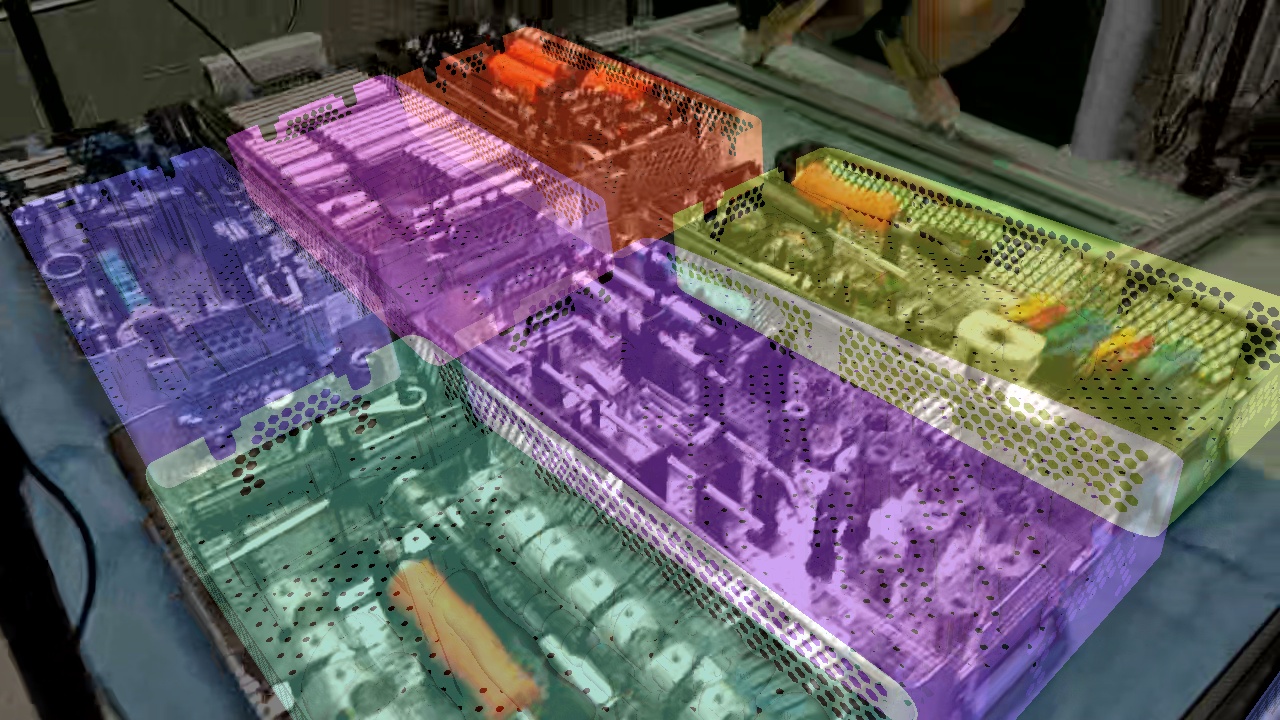}
                        &   \includegraphics[width=\hsize,valign=m]{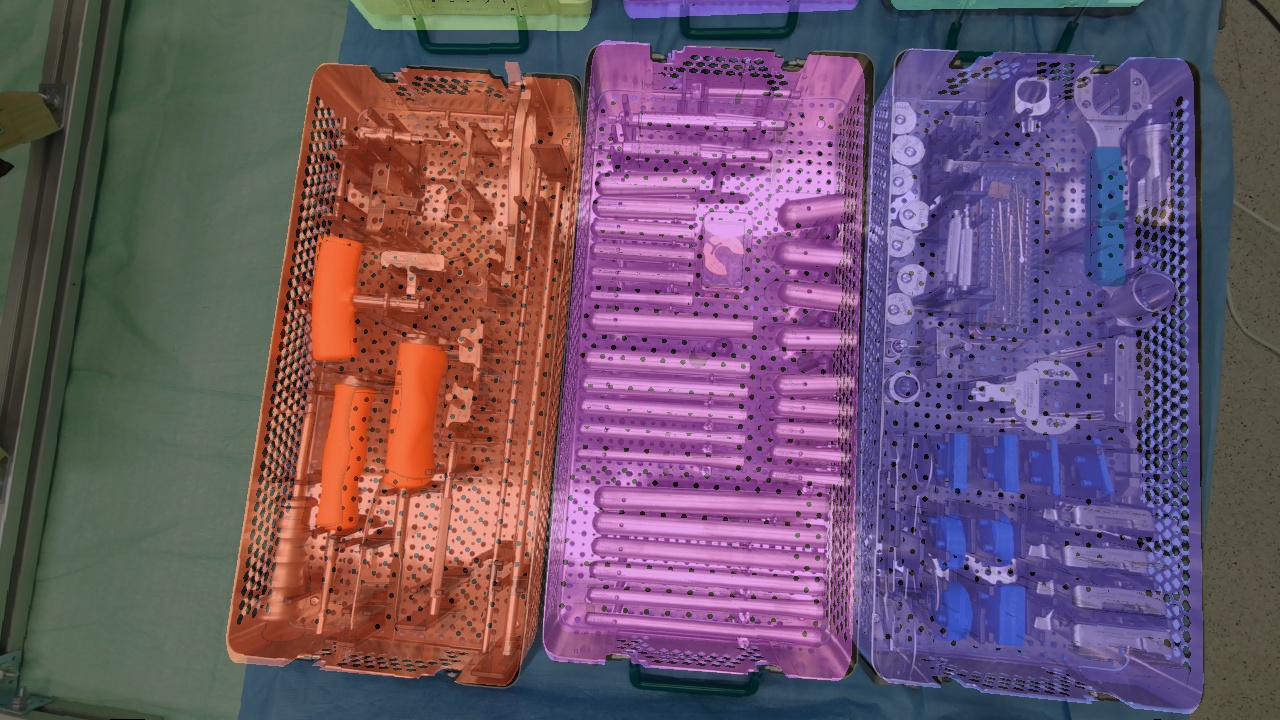}    
                        &   \includegraphics[width=\hsize,valign=m]{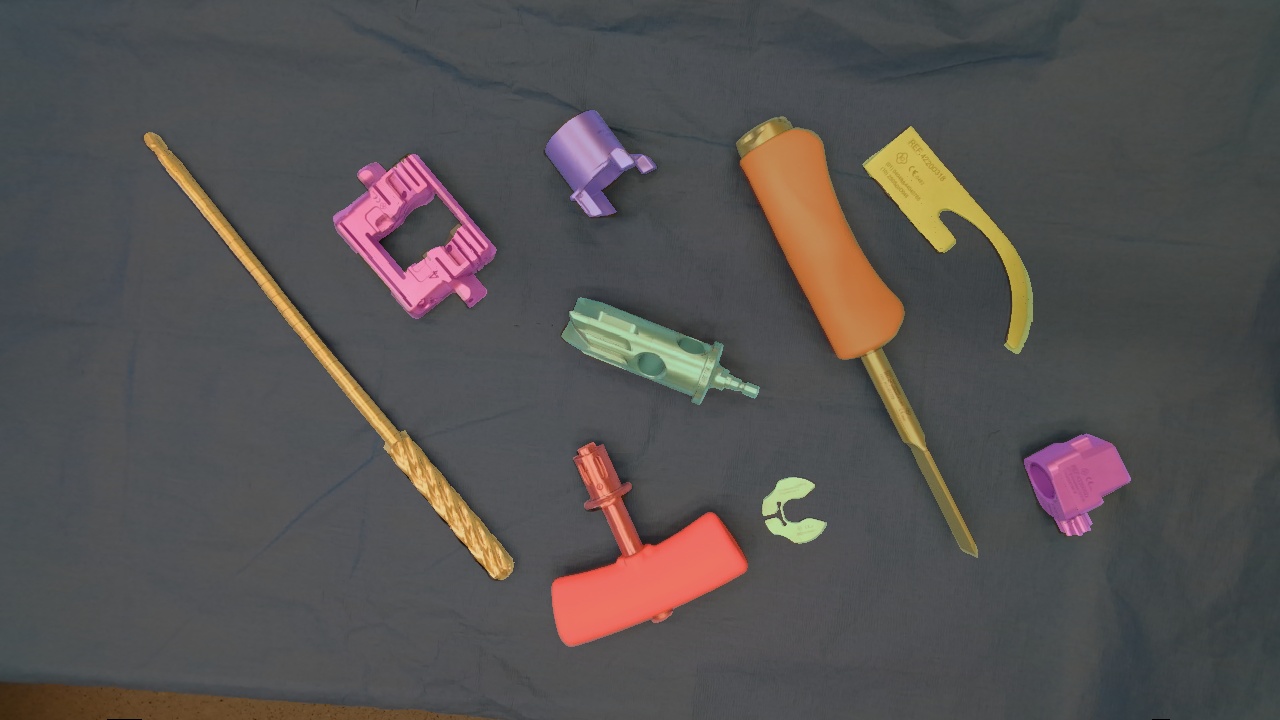}  
                        \\  \addlinespace[7pt]
                        
\rothead{CosyPose} &   \includegraphics[width=\hsize,valign=m]{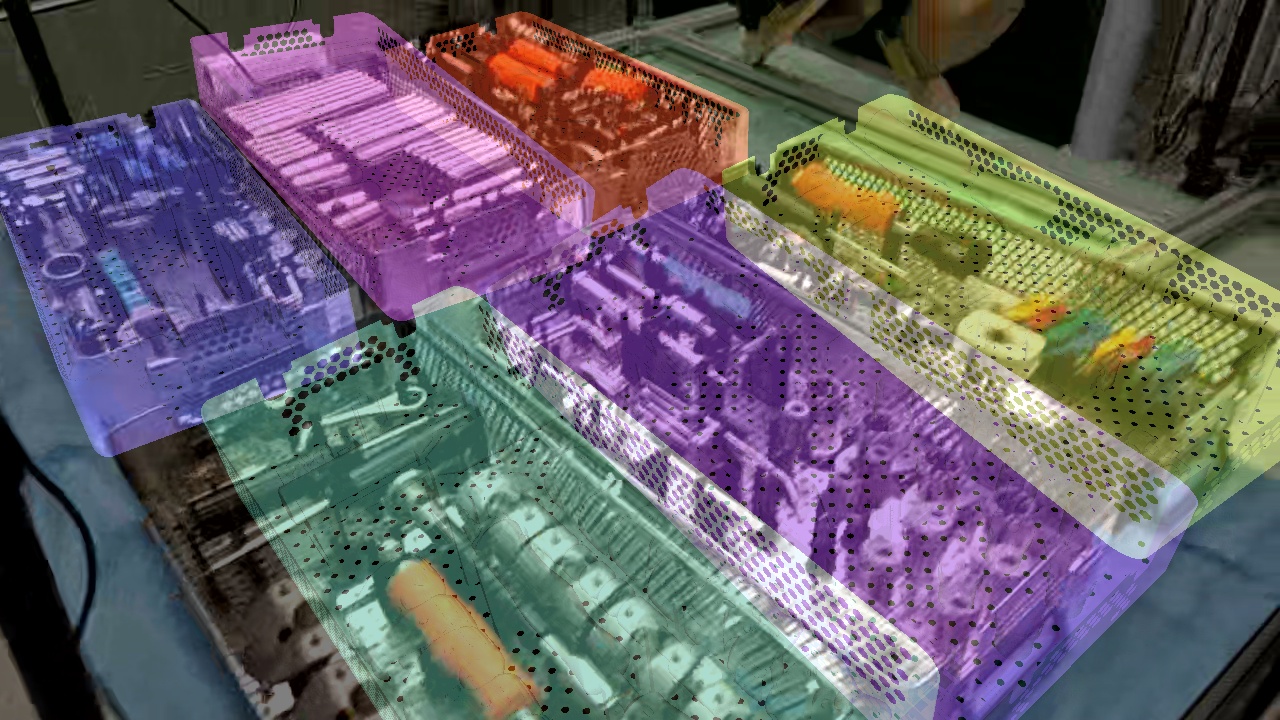}
                        &   \includegraphics[width=\hsize,valign=m]{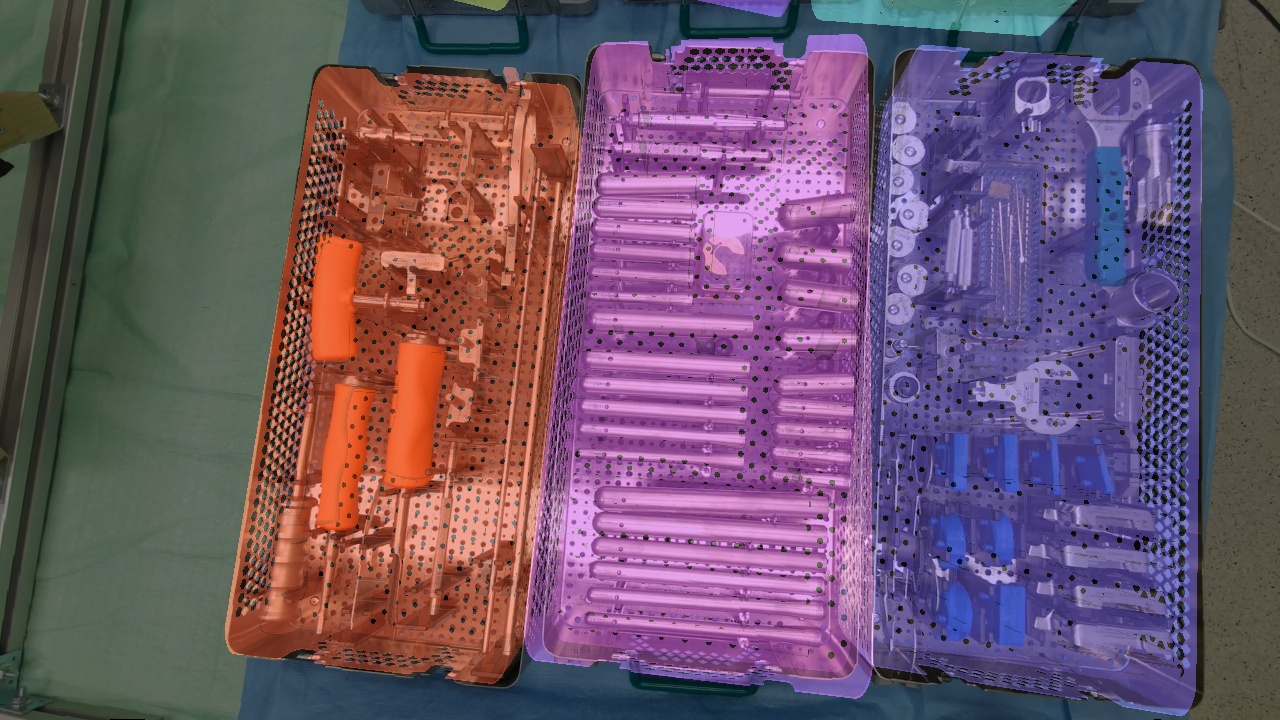}
                        &   \includegraphics[width=\hsize,valign=m]{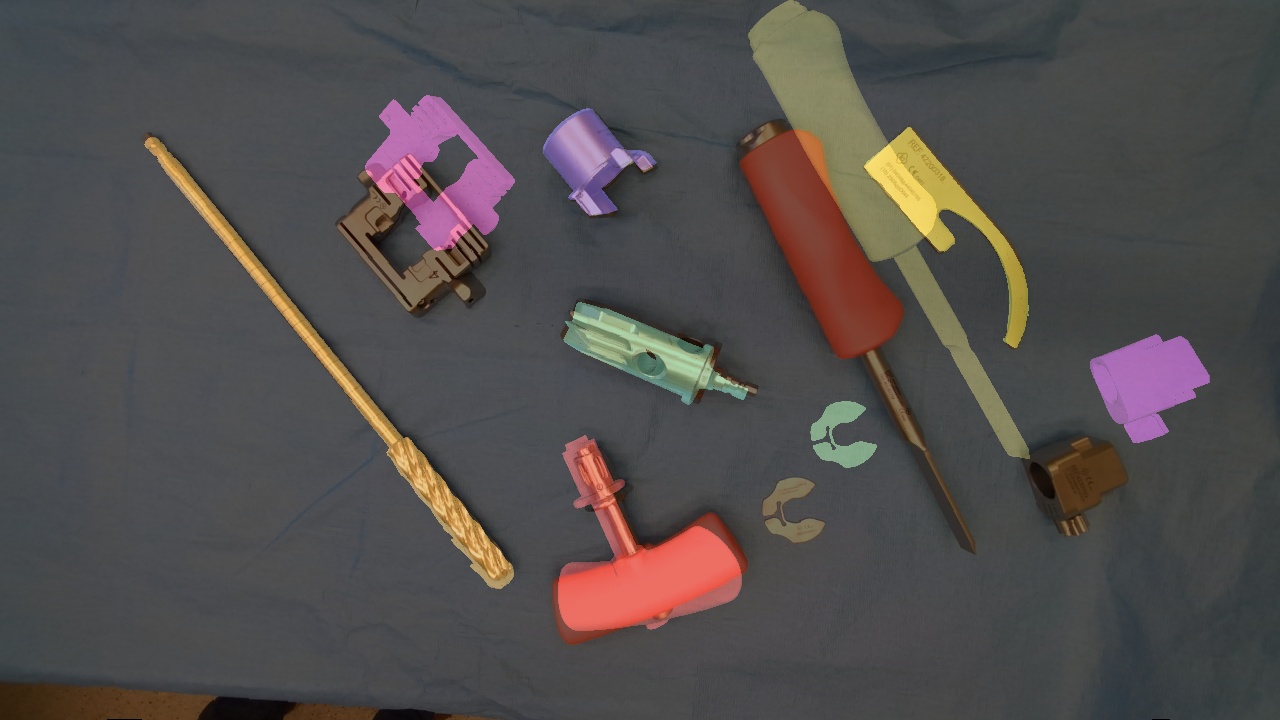}
                        \\  \addlinespace[7pt]

\rothead{Ours}        &   \includegraphics[width=\hsize,valign=m]{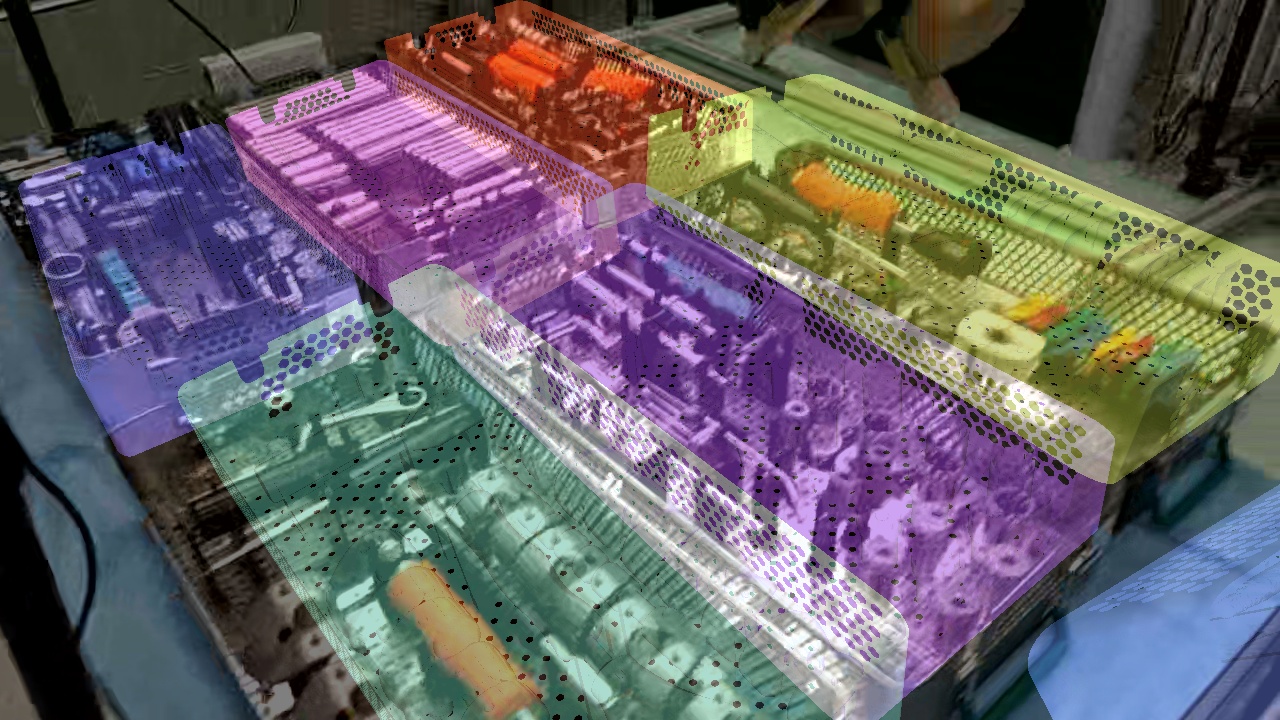}
                        &   \includegraphics[width=\hsize,valign=m]{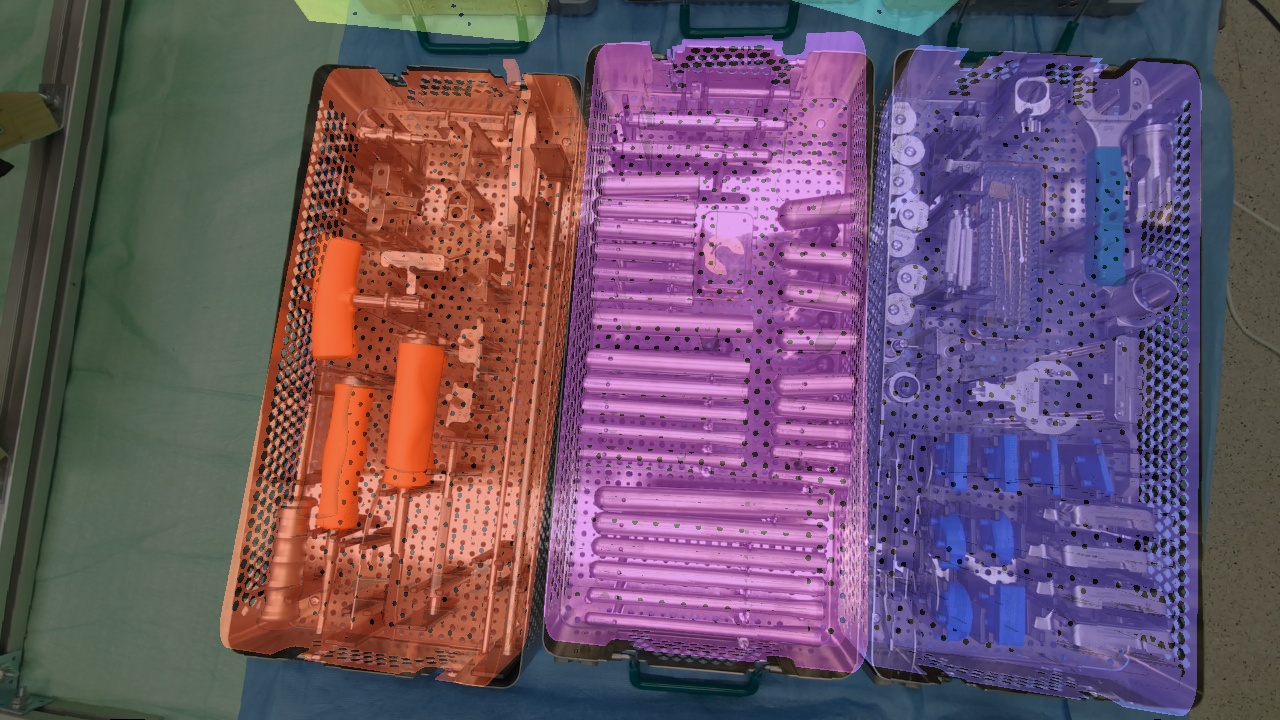}    
                        &   \includegraphics[width=\hsize,valign=m]{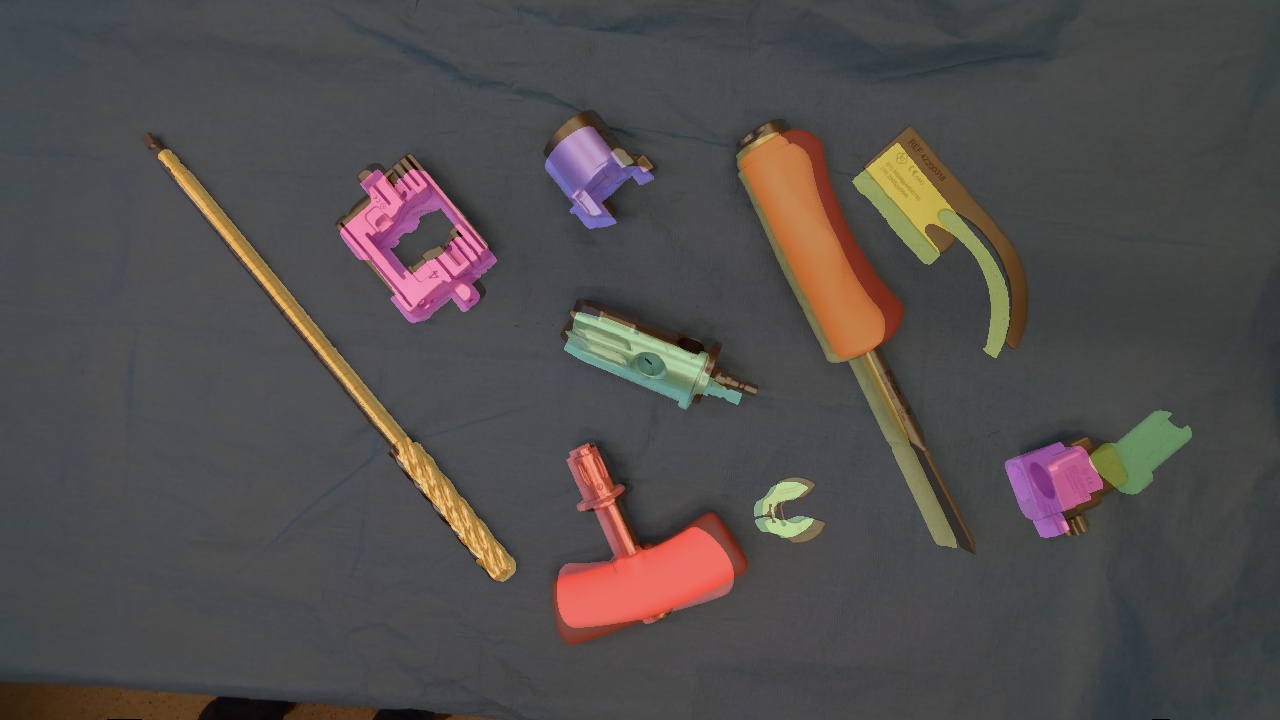}
                        \\  \addlinespace[7pt]
                        
\end{tabularx}
    \caption{Visualization of ground truth and pose estimation results for surgical trays and instruments. Our approach shows less visualization error compared to CosyPose \cite{labbe2020cosypose}.}
\label{fig:OP_multi}
\end{figure*}

To technically evaluate our system, we evaluate runtime and pose estimation accuracy in ADD(-S) using 3785 frames images in 720P resolution from one \ac{hmd} (HoloLens 2) and four static external RGB cameras (Azure Kinect). Our ground truth data is annotated by Optitrack motion capture system with Motive 3.1.4 and six infrared Primex 13 cameras. Furthermore, to assess the quality of the multi-view multi-object evaluation, we compare our method to CosyPose \cite{labbe2020cosypose}, as shown in \autoref{tab:op_quan_eval}. Our method shows overall better accuracy with $ADD(-S) AUC$ 44.34 for trays and 31.73 for instruments due to improved scene matching and pose filter. The camera pose error is less than 70 mm and rotation error less than 6 degrees. \autoref{fig:OP_multi} shows comparison of qualitative results.

\begin{table*}[t!]
\caption{\textbf{Camera and object pose evaluation on our test data.} 
The best results among all methods are labeled in bold.}
    \label {tab:op_quan_eval}
\centering
   \resizebox{\textwidth}{!}{    
        \begin{tabular}{l|c|c|c|cc|c}\toprule
        \multirow{2}{*}{Assets} &
        \multirow{2}{*}{Methods} & 
        \multicolumn{1}{c|}{{\centering Condition}}&

        \multicolumn{1}{c|}{{\centering Object pose}}&
        \multicolumn{2}{c|}{{\centering Camera pose}}&
        Overall runtime\\ 

         &
         & pose filter
         & $ADD(-S) AUC \uparrow$
         & $e_{trans} \downarrow$  & $e_{ rot} \downarrow$
         & ms \\ \midrule
         \multirow{3}{3cm}{Entire trays}
             
             &\multirow{1}{3cm}{CosyPose~\cite{labbe2020cosypose}}
                   & \xmark  & 38.26 &84.46&9.79& \textbf{183.47}\\ \cmidrule{2-7}
             &\multirow{2}{3cm}{Ours}  &\xmark  & 21.80 & 73.81 & 7.12 &186.92\\  
                   &&\cmark & \textbf{44.34} & \textbf{66.95} & \textbf{5.20} & 184.95\\
               \midrule
        \multirow{3}{3cm}{Instruments in trays}
            &\multirow{1}{3cm}{CosyPose~\cite{labbe2020cosypose}}
                  & \xmark  & 16.94 &94.14&8.37&167.37\\ \cmidrule{2-7}
            &\multirow{2}{3cm}{Ours}  
                  &\xmark  & 24.24 & 72.56 & 6.13 & \textbf{111.59}\\
                  & &\cmark  & \textbf{31.73} & \textbf{69.66} & \textbf{5.79} & 118.08\\

        \bottomrule
        \end{tabular}}
\end{table*}

\section{User Study}
Our system was compared to a paper manual (current gold standard) in a within-subjects study using a simulated surgical scenario, see Fig.~\ref{fig:teaser}. We measured assembly time, error rate, and workload (raw NASA TLX) \cite{tubbs-cooley_nasa_2018}. Ethical approval was obtained from the ethics board of Technical University of Munich.

\subsubsection{Sample, Procedure, and Task}
We recruited 29 scrub nurses (25 female, 4 male; $30.9 \pm 10.9$ years; \ac{or} experience from 0.5 to 32 years) participated. 22 used the tested Implantcast GenuX MK implant $\leq3$ times (Fig. \ref{fig:boxesplots} mid shows the distribution of experience). First, participants completed an \ac{ar} tutorial using a subset of the TFNA instruments for familiarization. Second, they assisted two parts of a simulated knee arthroplasty surgery (tibia and femur preparation, each~25–30 parts and~15–40 min), comparing paper manuals and our \ac{ar} system. Counterbalancing was complete for 28 participants. An additional participant introduces only negligible imbalance without biasing condition comparisons.

\subsubsection{Quantitative Results}

\begin{figure}[t]
    \centering
   \includegraphics[trim= 0 0 0 0, clip, width=0.25\linewidth]{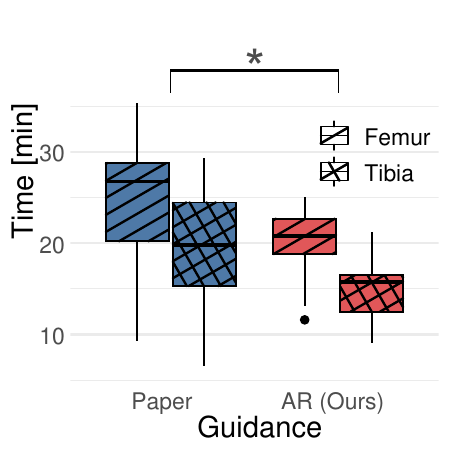}
   \includegraphics[trim= 0 0 0 40, clip, width=0.4\linewidth]{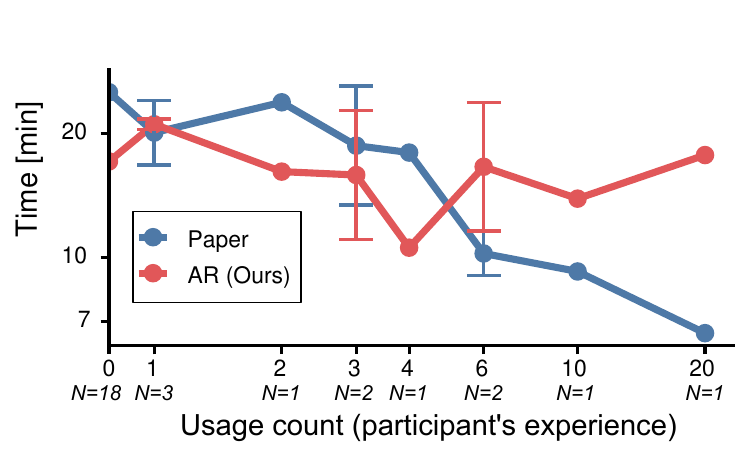}
   \includegraphics[trim= 0 0 0 0, clip, width=0.25\linewidth]{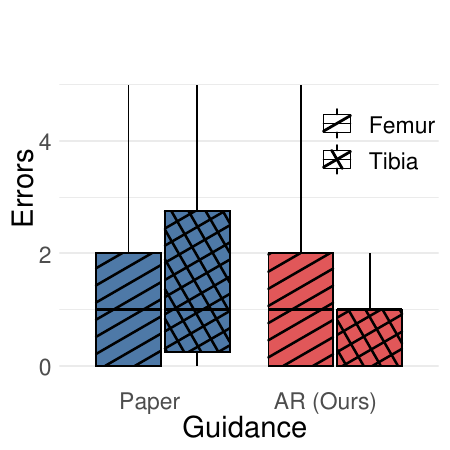}
    \caption{Boxplots of simulated surgery times for each guidance type and surgery part (left), participants experience (middle), and errors (right). Considering the experience, the results show that the system is beneficial for practitioners with fewer than 4 prior practical OR experiences with the sets.}
    \label{fig:boxesplots}
\end{figure}

Our system significantly outperformed the paper manual in time to completion ($F(1) = 12.036, p=.001, \eta_{p}^2=.108$), see \autoref{fig:boxesplots}. We found main effects for the type of bone ($F(1) = 16.989, p<.001, \eta_{p}^2=.167$) and experience with the instruments ($F(3) = 5.455, p=.003, \eta_{p}^2=.160$). This underlines that specifically less experienced staff benefit from the \ac{ar} guidance, whereas a turning point can be identified between 4 and 6 previous uses. With AR, also the timing differences between femur and tibia in terms of complexity become clearer than with paper. No significant differences were observed in the error rates. Our system further resulted in a significantly lower perceived workload (mental, temporal, performance, effort, and frustration; see \autoref{tab:results}).

\begin{table}[t]
    \centering
        \caption{Subjective results for paper and \ac{ar} condition.}
    \begin{tabular}{l|cc cc cc}  \toprule
          NASA-TLX  & $P_M$ & $P_{SD}$ & $AR_M$ & $AR_{SD}$ &  $p\leq$ & Effect\\ \hline 
        ~~ Mental & 75.86 & 17.88 & 43.28 &  26.33 & \textbf{.001} & $r=0.84$ \\ 
        ~~ Physical & 35.35 & 27.61 & 27.24 &  23.78 & .266 & $r=0.21$ \\
        ~~ Temporal & 73.28 & 20.63 & 33.45 &  24.28 & \textbf{.001} & $r=0.96$ \\
        ~~ Performance & 53.45 & 24.39 & 30.17 &  18.54 &  \textbf{.001} & $d=1.07$ \\ 
        ~~ Effort & 63.45 & 22.96 & 32.07  & 15.15 &  \textbf{.001} &  $d=1.59$ \\ 
        ~~ Frustration &  68.28 & 20.37 & 33.97 & 23.16 & \textbf{.001} &  $d=1.57$ \\
        \bottomrule    
    \end{tabular}
    \label{tab:results}
\end{table}

\subsubsection{Qualitative Results}

The participants expressed an overall positive attitude toward our approach due to the step-wise instructions, clear menu structure, and the combination of assembly animations with tooltips on parts to pick. In contrast to the paper manual, selective \ac{ar} information presentation was associated with effective visualization and reduced information overload. 
The system was described as particularly suitable for training and supporting users with limited prior experience. Both gaze and hand-based interaction (moveable user interface) were described as supportive of system use. 

\section{Discussion}

We introduce an \ac{ar} guidance system driven by real-time 6D pose estimation trained on synthetic data. We outperform CosyPose\cite{labbe2020cosypose} in evaluation metrics. Given manufacturers' CAD models, our approach could be scaled to manifold instrument sets.
We support a flexible, multi-component instrument setup. Our non-overlapping multi-camera setup is used for \ac{fov} extension, rather than a multi-view pose accuracy improvement \cite{labbe2020cosypose,hein2025next,agethen2025recurrent}. Our approach is leveraging known object knowledge \cite{li2026multicam} and combines hierarchical fusion and dynamic filtering. This enables dynamic camera calibration, and thus cameras/trays to be repositioned without manual recalibration. It is further extendable in expansion. To our knowledge, this is the first AR system for pick, assemble, and handover tasks for scrub nurses, rather than surgeons’ tasks \cite{yang2025augmented,acar2025navius}.

The present system was validated by scrub nurses in a simulated surgery, demonstrating a clear advantage over traditional paper-based instructions for instrumentation. As intended, our system is especially beneficial for inexperienced scrub nurses with a turnover point between 4 and 6 previous uses (Fig. \ref{fig:boxesplots}, middle). Error frequencies remained similar. Time differences for femur and tibia preparation become more apparent, as the guided handling of more complex parts scales. Overall, handling our \ac{ar} system was perceived as positive throughout.

\textbf{Limitations} that were commented on by the participants include ergonomic constraints, a limited field of view, sterility concerns, and acclimatization requirements (learning curve). Technical challenges included rare gaze calibration problems, delayed feedback, and tooltip inaccuracies. The fact that guidance text instructions have to be implemented manually represents a remaining challenge for the future.

\textbf{Future Work} will focus on improving pose estimation accuracy, adaptive and context-aware visualization, and validation under real \ac{or} conditions across additional instrument sets. Overall, this work establishes a baseline for integrating \ac{ar} guidance into routine surgical practice, potentially expanding to monitoring systems for experienced users.

\section{Conclusion}

We present a multi-view pose-estimation-driven \ac{ar} guidance system for surgical instrumentation, demonstrating measurable improvements in task performance and usability, particularly for inexperienced scrub nurses. By combining marker-free object pose estimation with spatially registered visualization, the system provides context-aware, situated guidance that supports efficiency, reduces uncertainty, and enhances procedural confidence in surgical workflows. The findings indicate that the guidance paradigm itself is technically robust and clinically promising.

\begin{credits}
\subsubsection{\ackname}

This work was funded by the German Federal Ministry of Research, Technology and Space (BMFTR) in the grant program "AI-based assistance systems for in-process health applications", grant number 16SV8973. Shiyu Li acknowledges the financial support from the China Scholarship Council. This work was partially supported by the TU Munich, through its MIRMI Seed Fund program, and by the Robotics Institute Germany (RIG) under grant 16ME0997K. The authors thank implantcast GmbH for lending the surgical equipment. We acknowledge HPC resources (NHR@FAU b143dc, b180dc), funded by federal and Bavarian state authorities, and Gerhard Wellein's and the NHR team's HPC approach. NHR@FAU hardware is partially funded by DFG 440719683. Additional support by ERC projects MIA-NORMAL 101083647, DFG 513220538 and 512819079, and the state of Bavaria (HTA and the Bavarian Foundation Model Initiative). 

\subsubsection{\discintname}
The authors have no relevant competing interests.
\end{credits}

%
%
%
\bibliographystyle{splncs04}
\bibliography{mybibliography}
%




\end{document}

%% file: acronym.tex
\begin{acronym}[Bspwwww.]  

\acro{ap}[AP]{average precision}
\acro{api}[API]{application programming interface}
\acroplural{ann}[ANN]{artifical neural networks}
\acro{bev}[BEV]{bird eye view}
\acro{bbox}[bbox]{bounding boxes}
\acro{rbob}[BRB]{Bottleneck residual block}
\acroplural{rbob}[BRBs]{Bottleneck residual blocks}
\acro{mbiou}[mBIoU]{mean Boundary Intersection over Union}
\acro{cai}[CAI]{computer-assisted intervention}
\acro{ce}[CE]{cross entropy}
\acro{cad}[CAD]{computer-aided design}
\acro{cnn}[CNN]{convolutional neural network}

\acro{crf}[CRF]{conditional random fields}
\acro{dpc}[DPC]{dense prediction cells}
\acro{dla}[DLA]{deep layer aggregation}
\acro{dnn}[DNN]{deep neural network}
\acroplural{dnn}[DNNs]{deep neural networks}

\acro{da}[DA]{domain adaption}
\acro{dr}[DR]{domain randomization}
\acro{fat}[FAT]{falling things}
\acro{fcn}[FCN]{fully convolutional network}
\acroplural{fcn}[FCNs]{fully convolutional networks}
\acro{fov}[FoV]{field of view}
\acro{fv}[FV]{front view}
\acro{fp}[FP]{False Positive}
\acro{fpn}[FPN]{feature Pyramid network}
\acro{fn}[FN]{False Negative}
\acro{fmss}[FMSS]{fast motion sickness scale}
\acro{gan}[GAN]{generative adversarial network}
\acroplural{gan}[GANs]{generative adversarial networks}
\acro{gcn}[GCN]{graph convolutional network}
\acroplural{gcn}[GCNs]{graph convolutional networks}
\acro{gs}[GS]{Gaussian Splatting}
\acro{hmi}[HMI]{Human-Machine-Interaction}
\acro{hmd}[HMD]{Head Mounted Display}
\acroplural{hmd}[HMDs]{head mounted displays}
\acro{iou}[IoU]{intersection over union}
\acro{irb}[IRB]{inverted residual bock}
\acroplural{irb}[IRBs]{inverted residual blocks}
\acro{ipq}[IPQ]{igroup presence questionnaire}
\acro{knn}[KNN]{k-nearest-neighbor}
\acro{lidar}[LiDAR]{light detection and ranging}
\acro{lsfe}[LSFE]{large scale feature extractor}
\acro{llm}[LLM]{large language model}
\acro{map}[mAP]{mean average precision}
\acro{mc}[MC]{mismatch correction module}
\acro{miou}[mIoU]{mean intersection over union}
\acro{mis}[MIS]{Minimally Invasive Surgery}
\acro{msdl}[MSDL]{Multi-Scale Dice Loss}
\acro{ml}[ML]{Machine Learning}
\acro{mlp}[MLP]{multilayer perception}
\acro{miou}[mIoU]{mean Intersection over Union}
\acro{nn}[NN]{neural network}
\acroplural{nn}[NNs]{neural networks}
\acro{ndd}[NDDS]{NVIDIA Deep Learning Data Synthesizer}
\acro{nocs}[NOCS]{Normalized Object Coordiante Space}
\acro{nerf}[NeRF]{Neural Radiance Fields}
\acro{NVISII}[NVISII]{NVIDIA Scene Imaging Interface}
\acro{ngp}[NGP]{Neural Graphic Primitives}
\acro{or}[OR]{Operating Room}
\acro{pbr}[PBR]{physically based rendering}
\acro{psnr}[PSNR]{peak signal-to-noise ratio}
\acro{pnp}[PnP]{Perspective-n-Point}
\acro{rv}[RV]{range view}
\acro{roi}[ROI]{region of interest}
\acroplural{roi}[ROIs]{region of interests}
\acro{rbab}[BB]{residual basic block}
\acro{ras}[RAS]{robot-assisted surgery}
\acroplural{rbab}[BBs]{residual basic blocks}
\acro{spp}[SPP]{spatial pyramid pooling}
\acro{sh}[SH]{spherical harmonics}
\acro{sgd}[SGD]{stochastic gradient descent}
\acro{sdf}[SDF]{signed distance function}
\acro{sfm}[SfM]{structure-from-motion}
\acro{sam}[SAM]{Segment-Anything}
\acro{sus}[SUS]{system usability scale}
\acro{ssim}[SSIM]{similarity index measure}
\acro{sfm}[SfM]{structure from motion}
\acro{slam}[SLAM]{simultaneous localization and mapping}
\acro{tp}[TP]{True Positive}
\acro{tn}[TN]{True Negative}
\acro{thor}[thor]{The House Of inteRactions}
\acro{tsdf}[TSDF]{signed distance function}
\acro{vr}[VR]{Virtual Reality}
\acro{ycb}[YCB]{Yale-CMU-Berkeley}

\acro{ar}[AR]{Augmented Reality}
\acro{ate}[ATE]{absolute trajectory error}
\acro{bvip}[BVIP]{blind or visually impaired people}
\acro{cnn}[CNN]{convolutional neural network}
\acro{c2f}[c2f]{coarse-to-fine}
\acro{fov}[FoV]{field of view}
\acro{gan}[GAN]{generative adversarial network}
\acro{gcn}[GCN]{graph convolutional Network}
\acro{gnn}[GNN]{Graph Neural Network}
\acro{hmi}[HMI]{Human-Machine-Interaction}
\acro{hmd}[HMD]{head-mounted display}
\acro{mr}[MR]{mixed reality}
\acro{iot}[IoT]{internet of things}
\acro{llff}[LLFF]{Local Light Field Fusion}
\acro{bleff}[BLEFF]{Blender Forward Facing}

\acro{lpips}[LPIPS]{learned perceptual image patch similarity}
\acro{nerf}[NeRF]{neural radiance fields}
\acro{nvs}[NVS]{novel view synthesis}
\acro{mlp}[MLP]{multilayer perceptron}
\acro{mrs}[MRS]{Mixed Region Sampling}

\acro{or}[OR]{Operating Room}
\acro{pbr}[PBR]{physically based rendering}
\acro{psnr}[PSNR]{peak signal-to-noise ratio}
\acro{pnp}[PnP]{Perspective-n-Point}
%
\acro{sus}[SUS]{system usability scale}
\acro{ssim}[SSIM]{similarity index measure}
\acro{sfm}[SfM]{structure from motion}
\acro{slam}[SLAM]{simultaneous localization and mapping}

\acro{tp}[TP]{True Positive}
\acro{tn}[TN]{True Negative}
\acro{thor}[thor]{The House Of inteRactions}
\acro{ueq}[UEQ]{User Experience Questionnaire}
\acro{vr}[VR]{virtual reality}
\acro{who}[WHO]{World Health Organization}
\acro{xr}[XR]{extended reality}
\acro{ycb}[YCB]{Yale-CMU-Berkeley}
\acro{yolo}[YOLO]{you only look once}
\end{acronym}

%% file: mybibliography.bib
@article{agethen2025recurrent,
  title={Recurrent multi-view 6DoF pose estimation for marker-less surgical tool tracking},
  author={Agethen, Niklas and Rosskamp, Janis and Koller, Tom L and Klein, Jan and Zachmann, Gabriel},
  journal={International Journal of Computer Assisted Radiology and Surgery},
  pages={1--11},
  year={2025},
  publisher={Springer}
}

@inproceedings{hogenkamp20256d,
  title={6D Object Pose Tracking for Orthopedic Surgical Training Using Visual-Inertial Sensor Fusion},
  author={Hogenkamp, Maarten and Stauffer, Tobias and Lohmeyer, Quentin and Meboldt, Mirko},
  booktitle={International Conference on Medical Image Computing and Computer-Assisted Intervention},
  pages={13--23},
  year={2025},
  organization={Springer}
}

@article{demir2023deep,
  title={Deep learning in surgical workflow analysis: a review of phase and step recognition},
  author={Demir, Kubilay Can and Schieber, Hannah and Weise, Tobias and Roth, Daniel and May, Matthias and Maier, Andreas and Yang, Seung Hee},
  journal={IEEE Journal of Biomedical and Health Informatics},
  volume={27},
  number={11},
  pages={5405--5417},
  year={2023},
  publisher={IEEE}
}

@inproceedings{schieber_asdf_2024,
	title = {{ASDF}: {Assembly} {State} {Detection} {Utilizing} {Late} {Fusion} by {Integrating} {6D} {Pose} {Estimation}},
	shorttitle = {{ASDF}},
	urldate = {2024-12-13},
	booktitle = {2024 {IEEE} {International} {Symposium} on {Mixed} and {Augmented} {Reality} ({ISMAR})},
	author = {Schieber, Hannah and Li, Shiyu and Corell, Niklas and Beckerle, Philipp and Kreimeier, Julian and Roth, Daniel},
	month = oct,
	year = {2024},
	note = {ISSN: 2473-0726},
	pages = {190--199},
}

@inproceedings{li_gbot_2024,
	title = {{GBOT}: {Graph}-{Based} {3D} {Object} {Tracking} for {Augmented} {Reality}-{Assisted} {Assembly} {Guidance}},
	shorttitle = {{GBOT}},
	urldate = {2024-12-13},
	booktitle = {2024 {IEEE} {Conference} {Virtual} {Reality} and {3D} {User} {Interfaces} ({VR})},
	author = {Li, Shiyu and Schieber, Hannah and Corell, Niklas and Egger, Bernhard and Kreimeier, Julian and Roth, Daniel},
	month = mar,
	year = {2024},
	note = {ISSN: 2642-5254},
	pages = {513--523},
}

@inproceedings{kreimeier_visual_2024,
	title = {Visual {Guidance} for {Assembly} {Processes}},
	urldate = {2024-12-13},
	booktitle = {2024 {IEEE} {International} {Symposium} on {Mixed} and {Augmented} {Reality} {Adjunct} ({ISMAR}-{Adjunct})},
	author = {Kreimeier, Julian and Schieber, Hannah and Li, Shiyu and Martin-Gomez, Alejandro and Roth, Daniel},
	month = oct,
	year = {2024},
	note = {ISSN: 2771-1110},
	pages = {652--653},
}

@article{cramer_requirement_2024,
	title = {Requirement analysis for an {AI}-based {AR} assistance system for surgical tools in the operating room: stakeholder requirements and technical perspectives},
	journal = {International Journal of Computer Assisted Radiology and Surgery},
	author = {Cramer, E and Kucharski, AB and Kreimeier, J and Andreß, S and Li, S and Walk, C and Merkl, F and Högl, J and Wucherer, P and Stefan, P and {others}},
	year = {2024},
	note = {Publisher: Springer},
	pages = {1--10},
}

@article{schieber_indoor_2024,
	title = {Indoor synthetic data generation: {A} systematic review},
	issn = {1077-3142},
	journal = {Computer Vision and Image Understanding},
	author = {Schieber, Hannah and Demir, Kubilay Can and Kleinbeck, Constantin and Yang, Seung Hee and Roth, Daniel},
	year = {2024},
	pages = {103907},
}

@inproceedings{denninger_blenderproc_2020,
	title = {{BlenderProc}: {Reducing} the {Reality} {Gap} with {Photorealistic} {Rendering}},
	author = {Denninger, Maximilian and Sundermeyer, Martin and Winkelbauer, Dominik and Olefir, D. and Hodan, Tomas and Zidan, Youssef and Elbadrawy, Mohamad and Knauer, M. and Katam, Harinandan and Lodhi, A.},
	year = {2020},
}

@article{goras_tasks_2019,
	title = {Tasks, multitasking and interruptions among the surgical team in an operating room: a prospective observational study},
	volume = {9},
	issn = {2044-6055},
	shorttitle = {Tasks, multitasking and interruptions among the surgical team in an operating room},
	language = {eng},
	number = {5},
	journal = {BMJ open},
	author = {Göras, Camilla and Olin, Karolina and Unbeck, Maria and Pukk-Härenstam, Karin and Ehrenberg, Anna and Tessma, Mesfin Kassaye and Nilsson, Ulrica and Ekstedt, Mirjam},
	month = may,
	year = {2019},
	pmid = {31097486},
	pmcid = {PMC6530509},
	pages = {e026410},
}

@misc{noauthor_global_nodate,
	title = {Global {Patient} {Safety} {Action} {Plan} 2021-2030},
    author = {World Health Organization (WHO)},
	url = {https://www.who.int/teams/integrated-health-services/patient-safety/policy/global-patient-safety-action-plan},
	language = {en},
	urldate = {2023-09-07},
}

@inproceedings{hu_wide-depth-range_2021,
	title = {Wide-depth-range 6d object pose estimation in space},
	pages = {15870--15879},
	booktitle = {Proceedings of the {IEEE}/{CVF} Conference on Computer Vision and Pattern Recognition},
	author = {Hu, Yinlin and Speierer, Sebastien and Jakob, Wenzel and Fua, Pascal and Salzmann, Mathieu},
	year = {2021},
}

@inproceedings{blattgerste_-situ_2018,
	address = {Corfu Greece},
	title = {In-{Situ} {Instructions} {Exceed} {Side}-by-{Side} {Instructions} in {Augmented} {Reality} {Assisted} {Assembly}},
	isbn = {978-1-4503-6390-7},
	language = {en},
	urldate = {2023-05-08},
	booktitle = {Proceedings of the 11th {PErvasive} {Technologies} {Related} to {Assistive} {Environments} {Conference}},
	publisher = {ACM},
	author = {Blattgerste, Jonas and Renner, Patrick and Strenge, Benjamin and Pfeiffer, Thies},
	month = jun,
	year = {2018},
	pages = {133--140},
}

@article{khuong_effectiveness_2014,
	series = {Proceedings - {IEEE} {Virtual} {Reality}},
	title = {The effectiveness of an {AR}-based context-aware assembly support system in object assembly: 21st {IEEE} {Virtual} {Reality} {Conference}, {VR} 2014},
	issn = {9781479928712},
	shorttitle = {The effectiveness of an {AR}-based context-aware assembly support system in object assembly},
	urldate = {2023-05-03},
	journal = {2014 IEEE Virtual Reality, VR 2014 - Proceedings},
	author = {Khuong, Bui Minh and Kiyokawa, Kiyoshi and Miller, Andrew and La Viola, Joseph J. and Mashita, Tomohiro and Takemura, Haruo},
	year = {2014},
	note = {Publisher: IEEE Computer Society},
	pages = {57--62},
}

@article{tubbs-cooley_nasa_2018,
	title = {The {NASA} {Task} {Load} {Index} as a measure of overall workload among neonatal, paediatric and adult intensive care nurses},
	volume = {46},
	issn = {0964-3397},
	language = {en},
	journal = {Intensive and Critical Care Nursing},
	author = {Tubbs-Cooley, Heather L. and Mara, Constance A. and Carle, Adam C. and Gurses, Ayse P.},
	month = jun,
	year = {2018},
	pages = {64--69},
}

@inproceedings{kleinbeck_artfm_2022,
	title = {{ARTFM}: {Augmented} {Reality} {Visualization} of {Tool} {Functionality} {Manuals} in {Operating} {Rooms}},
	shorttitle = {{ARTFM}},
	booktitle = {2022 {IEEE} {Conference} on {Virtual} {Reality} and {3D} {User} {Interfaces} {Abstracts} and {Workshops} ({VRW})},
	author = {Kleinbeck, Constantin and Schieber, Hannah and Andress, Sebastian and Krautz, Christian and Roth, Daniel},
	month = mar,
	year = {2022},
	pages = {736--737},
}

@article{dibene2022hololens,
  title={HoloLens 2 Sensor Streaming},
  author={Dibene, Juan C and Dunn, Enrique},
  journal={arXiv preprint arXiv:2211.02648},
  year={2022}
}

@misc{lu2023rtmo,
      title={{RTMO}: Towards High-Performance One-Stage Real-Time Multi-Person Pose Estimation},
      author={Peng Lu and Tao Jiang and Yining Li and Xiangtai Li and Kai Chen and Wenming Yang},
      year={2023},
      eprint={2312.07526},
      archivePrefix={arXiv},
      primaryClass={cs.CV}
}

@misc{mmpose2020,
    title={OpenMMLab Pose Estimation Toolbox and Benchmark},
    author={MMPose Contributors},
    howpublished = {\url{https://github.com/open-mmlab/mmpose}},
    year={2020}
}

@article{yolox2021,
  title={YOLOX: Exceeding YOLO Series in 2021},
  author={Ge, Zheng and Liu, Songtao and Wang, Feng and Li, Zeming and Sun, Jian},
  journal={arXiv preprint arXiv:2107.08430},
  year={2021}
}

@inproceedings{labbe2020cosypose,
  title={Cosypose: Consistent multi-view multi-object 6d pose estimation},
  author={Labb{\'e}, Yann and Carpentier, Justin and Aubry, Mathieu and Sivic, Josef},
  booktitle={European Conference on Computer Vision},
  pages={574--591},
  year={2020},
  organization={Springer}
}

@article{li2026multicam,
  title={MultiCam: On-the-fly Multi-Camera Pose Estimation Using Spatiotemporal Overlaps of Known Objects},
  author={Li, Shiyu and Schieber, Hannah and Waldow, Kristoffer and Busam, Benjamin and Kreimeier, Julian and Roth, Daniel},
  journal={IEEE Transactions on Visualization and Computer Graphics},
  year={2026},
  publisher={IEEE}
}

@inproceedings{yang2025augmented,
  title={Augmented Reality-Based Guidance with Deformable Registration in Head and Neck Tumor Resection},
  author={Yang, Qingyun and Li, Fangjie and Xu, Jiayi and Liu, Zixuan and Sridhar, Sindhura and Jin, Whitney and Du, Jennifer and Heiselman, Jon and Miga, Michael and Topf, Michael and others},
  booktitle={International Conference on Medical Image Computing and Computer-Assisted Intervention},
  pages={45--54},
  year={2025},
  organization={Springer}
}

@inproceedings{acar2025navius,
  title={NAVIUS: navigated augmented reality visualization for ureteroscopic surgery},
  author={Acar, Ayberk and Atoum, Jumanh and Connor, Peter S and Pierre, Clifford and Lynch, Carisa N and Kavoussi, Nicholas L and Wu, Jie Ying},
  booktitle={International Conference on Medical Image Computing and Computer-Assisted Intervention},
  pages={433--443},
  year={2025},
  organization={Springer}
}

@article{hein2025next,
  title={Next-generation surgical navigation: Marker-less multi-view 6DoF pose estimation of surgical instruments},
  author={Hein, Jonas and Cavalcanti, Nicola and Suter, Daniel and Zingg, Lukas and Carrillo, Fabio and Calvet, Lilian and Farshad, Mazda and Navab, Nassir and Pollefeys, Marc and F{\"u}rnstahl, Philipp},
  journal={Medical Image Analysis},
  volume={103},
  pages={103613},
  year={2025},
  publisher={Elsevier}
}

@article{muller2020augmented,
  title={Augmented reality navigation for spinal pedicle screw instrumentation using intraoperative 3D imaging},
  author={M{\"u}ller, Fabio and Roner, Simon and Liebmann, Florentin and Spirig, Jos{\'e} M and F{\"u}rnstahl, Philipp and Farshad, Mazda},
  journal={The Spine Journal},
  volume={20},
  number={4},
  pages={621--628},
  year={2020},
  publisher={Elsevier}
}

@article{martin2023sttar,
  title={STTAR: surgical tool tracking using off-the-shelf augmented reality head-mounted displays},
  author={Martin-Gomez, Alejandro and Li, Haowei and Song, Tianyu and Yang, Sheng and Wang, Guangzhi and Ding, Hui and Navab, Nassir and Zhao, Zhe and Armand, Mehran},
  journal={IEEE Transactions on Visualization and Computer Graphics},
  volume={30},
  number={7},
  pages={3578--3593},
  year={2023},
  publisher={IEEE}
}

@article{liebmann2024automatic,
  title={Automatic registration with continuous pose updates for marker-less surgical navigation in spine surgery},
  author={Liebmann, Florentin and von Atzigen, Marco and St{\"u}tz, Dominik and Wolf, Julian and Zingg, Lukas and Suter, Daniel and Cavalcanti, Nicola A and Leoty, Laura and Esfandiari, Hooman and Snedeker, Jess G and others},
  journal={Medical Image Analysis},
  volume={91},
  pages={103027},
  year={2024},
  publisher={Elsevier}
}
